\pdfoutput=1
\documentclass[11pt, conference]{IEEEtran}
\usepackage[english]{babel}
\usepackage[ruled,vlined]{algorithm2e}
\usepackage{graphicx}
\usepackage{hyperref}
\usepackage{bbm}
\hypersetup{
    colorlinks=true,
    linkcolor=blue,
    filecolor=magenta,      
    urlcolor=cyan,
    pdftitle={Sharelatex Example},
    pdfpagemode=FullScreen,
}
\begin{document}

\title{Reinforcement Learning For Data Poisoning on Graph Neural Networks}
\author{Jacob Dineen, \quad A S M Ahsan-Ul Haque, \quad Matthew Bielskas\\
Department of Computer Science, University of Virginia, Charlottesville, VA 22904\\
{\tt\small [jd5ed, ah3wj, mb6xn]@virginia.edu}
}

\maketitle

\begin{abstract}
  Adversarial Machine Learning has emerged as a substantial subfield of Computer Science due to a lack of robustness in the models we train along with crowdsourcing practices that enable attackers to tamper with data. In the last two years, interest has surged in adversarial attacks on graphs yet the Graph Classification setting remains nearly untouched. Since a Graph Classification dataset consists of discrete graphs with class labels, related work has forgone direct gradient optimization in favor of an indirect Reinforcement Learning approach. We will study the novel problem of Data Poisoning (training-time) attack on Neural Networks for Graph Classification using Reinforcement Learning Agents. 
\end{abstract}

\maketitle

\section{Introduction}
The success of Deep Learning algorithms in applications to tasks in computer vision, natural language processing, and reinforcement learning has induced cautious deployment behavior, particularly in safety-critical applications, due to their ability to be influenced by adversarial examples, or attacks \cite{jin2020adversarial, Xu2020AdversarialAA}. For real-world deployment, a level of trust in the robustness of algorithms to adversarial attacks is not only desired but required. Research by Goodfellow et al. in the areas of Generative Adversarial Networks (GANs) and Adversarial Training \cite{goodfellow2014generative, goodfellow2015explaining} has influenced a sub-field of Deep Learning focused on exactly the above, where the synthesis of viable attacks on neural network architectures, under White-Box, Black-Box or Grey-Box Attacks, can provide meaningful insight into both the models themselves and the preventative measures to be taken to ensure correctness (Adversarial Defenses \cite{Ren_Zheng_Qin_Liu_2020}). 

Here, we study a less saturated version of the adversarial setting: The effect of inducing a Poison attack on a Graph Neural Network under the task of graph classification \cite{Knyazev_Lin_Amer_Taylor_2018, Ying_You_Morris_Ren_Hamilton_Leskovec_2019}. The agent is trained to find an optimal policy, under Reinforcement Learning (RL) principles, such that they can inject, modify, or delete graph structure, or features under Black-Box Attacks. Such attacks assume that the agent has little to no underlying knowledge of the information regarding the target neural network (architecture, parameters, etc.), as is most often the case in real-world applications of machine learning algorithms, although ML-as-a-service systems provide infiltrators with alternative methods to attack \cite{tramer2016stealing}. We structure this problem as a Markov Decision Process, in which the agent must interact with their environment(s) (a set of graphs) to fool the underlying learning algorithm into graph misclassification.
\section{Background}

\subsection{Preliminaries \label{section: sec}}
We introduce, with brevity, necessary background information regarding concepts and notation spanning Graph Mining and Reinforcement Learning in this section.
\subsubsection{\textbf{Adversarial Attacks}}

In this paper, we consider an adversarial example to be an intentionally perturbed instance of data, whose purpose is to fool the learning algorithm \cite{10.1145/3374217}. The attacker's main goal is to induce model misspecification given a particular task such that imposed modifications to the original data are unbeknownst to the model or a human observer. Perturbations in the context of Graph Machine Learning are generally of the form of structure modification, i.e. node, edge, or sub-graph modification, or feature modifications concerning particular nodes in the graph. Such attacks fall under a taxonomy noted in \cite{Xu2020AdversarialAA}: Evasion attacks and Poisoning attacks. While our focus here will be on the latter, we introduce Evasion Attacks for completeness.

\textbf{Evasion attacks:} Evasion attacks are adversarial attacks where the underlying attack occurs after the specified learning algorithm is fully trained, e.g., the architecture and the learnable parameters are fixed and immutable.

\textbf{Poison attacks:} Poison attacks occur before or during the model training phase. In this way, we induce model misspecification during the parameter estimation phase of our learning paradigm.

\subsubsection{\textbf{Reinforcement Learning}}
The purpose of our work is to solve a Markov Decision Process (MDP) in which the agent(s) of the system is to solve sequential decision-making processes. MDPs are popular frameworks for superposing agent-environment interaction and simulation. Formally, an MDP can be represented by the tuple $\left(S, A, P_{a}, R_{a}\right)$, where $S$ is the state space, $A$ is the action space, $P_{a}$, or $T$ defines the dynamics of the state space (the probability that an action at a given state will lead to a subsequent state, the transition probability function), and $R_{a}$ is the reward function. Given the introduced context, attacker(s) of our learning algorithm is considered as agent(s). We establish common notation \cite{Sutton1998} but extend the basic definition of a Markov Decision Process via augmentation (Universal Markov Decision Processes).

\subsubsection{\textbf{Graph Convolutional Neural Networks (GNN)}}

GCNN is based on Convolutional Neural Network and graph embedding. Graph Neural Networks embed information from graph structures~\cite{zhou2018graph}. A GNN is essentially a CNN where the input is a graph (usually in the adjacency matrix form) and the feature vectors of every node of that graph.

In GNNs every node is represented by their  feature vectors. Then Neighborhood aggregation is performed for each node. Summing over the embedding of the nodes gives representation of the graph. An important application of GNN is node classification. For Node Classification, the GNN maps this input to the output feature vector of every node~\cite{kipf}. DeepWalk \cite{perozzi2014deepwalk} is such an algorithm for unsupervised node embedding. 

\subsubsection{\textbf{Graph Classification}}
Graph classification is the task of taking an instance of a graph and predicting its label(s). GNNs in this setting need to map node outputs to a graph-level output ("readout layer"), and then typically a Multi-Layer Perceptron (MLP) is employed to assign graph embeddings to class labels. \cite{tixier2019graph} have used 2D CNN  for Reddit, Collab and IMDB datasets in a supervised setting. \cite{kipf2016semi} have used GNN in a semi-supervised setting for Citeseer, Cora, Pubmed and NELL datasets. Formally, Graph Classification is employed over a set of attributed graphs, $\mathcal{D}=\left\{\left(G_{1}, \ell_{1}\right),\left(G_{2}, \ell_{2}\right), \cdots,\left(G_{n}, \ell_{n}\right)\right\}$, each containing a graph structure and a label, and learning a function $f: G \rightarrow \mathcal{L}$ where $G$ is the input space of graphs and $\mathcal{L}$ is the set of graph labels.

\subsubsection{\textbf{Notation}}
Each graph $G_{i} \in G$ is comprised of $V$ vertices, and $E$ edges. The cardinality of $V$ and $E$ corresponds to the size and the connectedness (edges) of the graph, respectively. We let $d^{i n}(G, i)$ represent the in-degree of the ith node of $G$.

\subsection{Related Work}
Attacks (on GNNs) in the Graph Classification setting are unique in that so far, they do not take the form of a gradient or meta-learning optimization problem directly on the model hyperparameters. The first attack introduced by \cite{Dai_Li_Tian_Huang_Wang_Zhu_Song_2018} is RL-S2V, a Hierarchical Reinforcement Learning (HRL) approach for both Node and Graph Classification at test-time. They propose this because HRL agents are better suited for attacking graph data which is both discrete and combinatorial. The authors formulate a Hierarchical Q Learning algorithm that relies on GNN parameterization, and they successfully train it to attack by adding or removing edges in test graphs. Ultimately this paper treats Graph Classification as an afterthought and does not provide an insightful way to alter RL-S2V in this setting.

An evasion attack designed for Graph Classification is Rewatt (\cite{Ma_Wang_Derr_Wu_Tang_2019}), which relies on standard Actor-Critic to perturb individual graphs. This is done via "rewiring" operations that simultaneously delete and add edges between nodes near each other. Rewiring is meant to be subtle so that graph instances maintain similar metrics such as degree centrality. The authors theoretically justify Rewatt by showing that rewiring preserves top eigenvalues in the graph's Laplacian matrix. While Rewatt is useful at test time, it is ill-suited for a Poisoning attack. This is because it is designed to attack a single graph with minimal perturbation. Meanwhile, we have the burden of selecting training graphs that would be best for poisoning, but we can get away with significantly altering graphs if we limit ourselves to a small portion of the training data.

The newest entry to the Graph Classification literature details a backdoor attack on Graph Neural Networks (\cite{Zhang_Jia_Wang_Gong_2020}). The purpose of a backdoor attack is to perturb data points in a particular manner (e.g. wearing an accessory in security camera images) so that they are mislabeled as a class of your choice. For Graph Classification, this is done by generating a random subgraph as a trigger and injecting it a portion of training graphs. The reasoning is that random subgraph generation (with good parameters) can produce graphs that aren't "too anomalous" yet are assigned to the attacker's class of choice by the tuned GNN. This paper is interesting because they manipulate subgraphs instead of edges; an RL agent that perturbs subgraphs has potential to learn quickly due to the reduction in its action space.

\section{Motivation}
Attacks on algorithms designed for classification/regression, detection, generative models, recurrent neural networks and deep reinforcement learning, and even graph neural networks, are well covered in literature \cite{papernot2016crafting, Akhtar_Mian_2018, lin2019tactics, jin2020adversarial}. The
generation of adversarial attacks in these spaces generally leads to corresponding defenses, such as Adversarial Training, Perturbation Detection \cite{xu2019characterizing}, and Graph Attention Mechanisms \cite{10.1145/3292500.3330851} among others. Uncovering the root cause of the model malfunction is an equally important part of the process towards widespread industrial adoption of Deep Neural Networks (DNNs).

As a practical example, we consider Wikipedia as a graphical representation - a network. Each node in the overarching network, composed of sub-graphs, could represent an article. Under the guise of a graph machine learning problem, a certain task for the learning algorithm could be to classify nodes as being 'real' or 'fake', e.g. "Is this article, that is linked to these other authentic articles, malicious?". A viable Graph Machine Learning algorithm would be able to detect or classify each observation to its correct class. A successful adversarial attack on the algorithm would be one that causes it, through perturbation of its underlying structure, to misclassify the instance. Extending this to a graph-level classification task, in chemoinformatics, molecules are represented as graph-structured objects \cite{Lee_Rossi_Kong_2018, Takigawa_Mamitsuka_2013}, and analysis, testing, and approval of chemical compounds can be detrimentally disturbed via attacks. Drug Discovery is a common application that directly impacts our well-being. Thus an adversary can thwart the automation of scientific progress if they can send training data e.g. through a crowdsourcing project. We see that many sociological and societal harms can result from misinformation propagation or malicious modification of data, which is why it's always important to expose vulnerabilities and defenses in deployed machine learning models.

To the best of our knowledge, we present a novel framework to the setting of adversarial data poisoning in graph classification tasks  \cite{Xu_Hu_Leskovec_Jegelka_2019}).

\section{Methodology}
The framework for initiating a Poison attack on a graph neural network classifier (GNN) can be decomposed into a multi-step process involving 1) Data Procurement, 2) Graph Classification, and 3) Reinforcement Learning. 
\begin{figure*}[!ht]
\centering
\includegraphics[width=13.5cm,height=6cm]{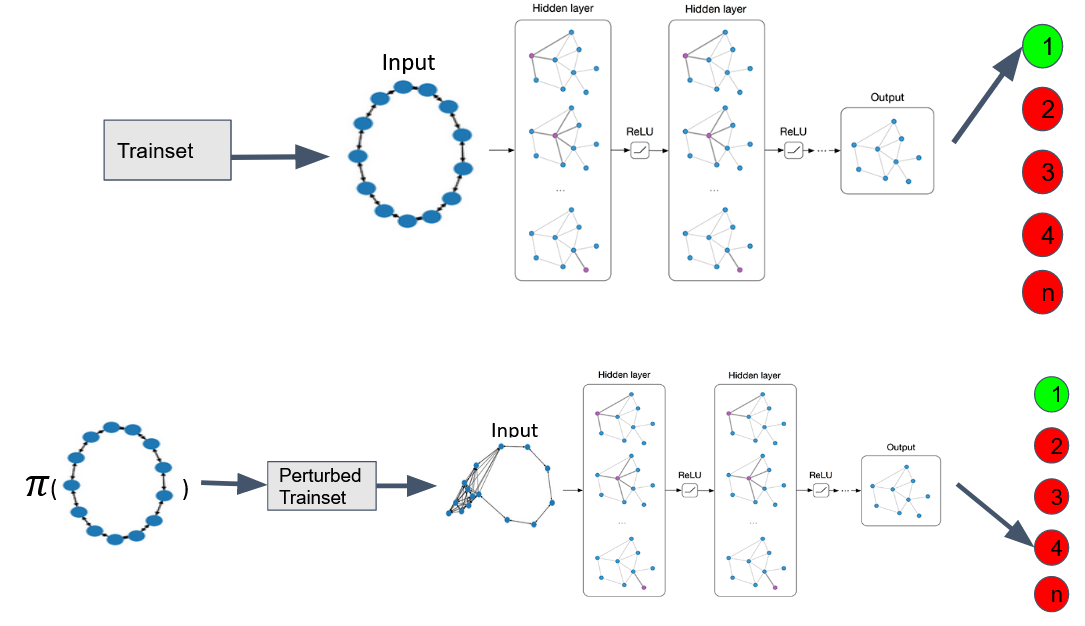}
\caption{Our Poison attack features a multi-step process, where the unperturbed model is used to generate a baseline generalization accuracy, and the RL algorithm is responsible for perturbing the original training data such that testing accuracy is degraded. Top: Running the original, unperturbed dataset through a GNN. Bottom: Applying REINFORCE to the training set to create a new dataset minimally perturbed by a threshold. Ideally, we see a case where these minimal perturbations lead to a large accuracy decrease come testing time.}
\label{fig:minigc}
\end{figure*}

\subsection{Data Procurement}
We require a graph dataset composed of a set of graph/label pairs. Each graph in the dataset is represented by a set of nodes and edges - $G(V, E)$. This dataset is instantiated or partitioned into two distinct sets, forming training and testing data. The training set has two purposes: 1) to train a GNN for the task of label classification over a set of graphs, and 2) to represent the state of the environment, whereby the reinforcement learning agent can act to perturb the state via a finite set of actions $A$. As noted in Section \ref{section: sec}, this is a Poison attack where we perturb the training set, rather than an evasion attack where the training data and GNN are untouched.

Experimenting with Graph Neural Networks has become more accessible with the release of \href{https://docs.dgl.ai/en/0.4.x/index.html}{Deep Graph Library} \cite{wang2019dgl}. It includes benchmark datasets, popular GNN algorithms across several settings, and easy compatibility with Pytorch and other Deep Learning libraries. Thus we will rely on DGL for graph models and data.

We will first experiment with a toy dataset class that DGL includes for Graph Classification. It features eight graph types such as circles and cubes. DGL also includes Graph Classification datasets that are baselines for Graph Kernel algorithms, such as PROTEINS, but our main focus of this work will be on the synthetic dataset.

\subsection{Graph Classification}\label{section:GNN}
The GNN, $f$, is a mapping from a raw graph $g$ to a discrete label associated with the respective class of the graph. We visualize the MiniGCDataset from DeepGraphLibary in Figure \ref{fig:minigc} to conceptualize the task of graph classification. As a comprehensive survey of GNNs is beyond the scope of this work, we use a simple multi-layer Graph Convolutional Neural Network (GNN) for all experiments which are described below.  

\subsection{Graph Neural Networks}
Here, $f$ features two convolutional layers, Relu activation functions, a readout function, and a fully connected linear layer. We use the node in degrees of $G \in G$, $x_{G}=d^{i n}(i, G),$ as input into the GNN and fix the readout function to $h_{g}=\frac{1}{|\mathcal{Y}|} \sum_{v \in \mathcal{V}} h_{v},$ averaging over node features for each graph in the batch. A batch of graphs $G \in V,$ where $V \subset G$ are fed through the GNN, where a graph representation $h_g$ is learned through message passing and graph convolutions over all nodes before being passed through to the linear layer of the network for classification \cite{wang2019dgl}. As most graph classification tasks that we explore with are multi-class, the outputs of the linear layer are passed through a softmax activation function: $\sigma(\mathbf{z})_{i}=\frac{e^{z_{i}}}{\sum_{j=1}^{K} e^{z_{j}}}$. 

\begin{figure}[!h]
\centering
\includegraphics[scale=.65]{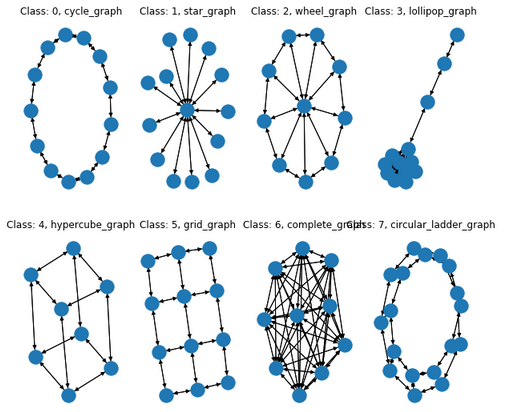}
\caption{Toy Classification Task: MiniGCDataset from DGL. Each attributed item is a graph, label pair. The graph generation process is parameterized by $MinNodes, MaxNodes$, and the probability of each class is uniform.}
\label{fig:poison}
\end{figure}
 
We employ categorical cross-entropy as our loss function $J$ throughout, and iteratively learn $\theta$ via backpropagation, seen in Equation \ref{equation: eq}.

\begin{equation}
\begin{array}{c}
\Delta w_{i j}=-\eta \frac{\partial J}{\partial w_{i j}} \\
w_{i j} \leftarrow w_{i j}-\eta \frac{\partial J}{\partial w_{i j}}
\end{array}
\label{equation: eq}
\end{equation}

We make note that we treat the hyper-parameterization of the GNN as fixed to reduce the elasticity of our experiments. Finely-tuned, and extensively trained neural networks may be less prone to poison attacks.

\subsection{Reinforcement Learning}
\subsubsection{\textbf{Algorithm}}
We use a Monte Carlo variant of a Deep Policy Gradient algorithm: \href{https://github.com/pytorch/examples/blob/master/reinforcement_learning/reinforce.py}{REINFORCE} \cite{Williams_1992}, with additional algorithmic validation saved for future work and use a simple random policy $\pi(s) = \frac{1}{|A|}$ as our baseline. Intuitively, we conceptualize this as: How much better does a Deep Reinforcement Learning Algorithm choose a sequence of poison points (graphs), such that testing accuracy is degraded, against a fully random search of the action space? 




\subsubsection{\textbf{Reward}}
We craft a custom Reward function, Equation \ref{equa : reward}, to pass through p (poison points) rewards during a single episode. Let $\theta$ be the original, learned parameters from training on the unperturbed set, and $\theta^p$ be the perturbed variant. $X^{test}$ is the original testing set generated a priori. For simplicity, we assume that the application of the actual or perturbed parameters to the testing set yields an accuracy. The Reward function $R$ is a signal that measures the difference between the unperturbed model's application to $X^{test}$ against the perturbed model's application to $X^{test}$. In other words, if the reward is positive, it means that the Poison attack was successful.

\begin{equation}
R = \theta(X) - \theta^p(X)
\label{equa : reward}
\end{equation}

\subsubsection{Environment}
Utilizing a form of Deep Reinforcement Learning, a set of inputs, representing the state of the system, are required at $t$ to pass through to a function approximator. We represent this state $S\left(G_{t}\right)$ as the mean, max, min over all $G_{i} \in G$ at $t$. Intuitively, the RL algorithm sees as input a tensor containing each graph's representation (defined by summary statistics) in the training set, chooses an action, and transitions into a new state based on the perturbations to the training set.

We define the action space $A$ in Table \ref{tab:table-name}. At any $t$, the agent can alter the state by performing any $a \in A$. To discretize the action space, we specify that the action of adding a subgraph into an existing graph's structure has a fixed set of parameters $n$ and $p$, corresponding to the number of nodes, and the probability of an edge between nodes, respectively. We also constrain the action space by permitting node or edge modifications to be random processes, e.g, the agent has no control over which node or edge to perturb, only that they wish to perform that action.

\begin{table}[!ht]
 \begin{tabular}{|p{1cm}||p{2cm}||p{4cm}|}
 \hline
 Action & Name & Definition  \\ [0.5ex] 
 \hline\hline
 $a_1$ & subgraph add & A random gnp graph is inserted into the existing graph structure. We fix $n = 10$ and $p = 0.75$ for fairness. \\ 
 \hline
\end{tabular}
\caption{\label{tab:table-name}Enumerating the action space of our environment.}
\end{table}

An optimal policy would learn which sequence of actions to take, and their applications to a subset of $G$, to maximize the episodic degradation in the GNN's ability to generalize over the testing set.

\subsection{Poison attack}
The culmination of the multi-step framework results in a poisoning attack on the original graph training set, re: Appendix \ref{algo : 1}. Given a set of graph, label pairs, we first compute a benchmark accuracy via the application of $\theta$ to the training set and perform inference on the testing set. Given a scalar value for $p$ representing the number of poison points that are selected by the RL algorithm at each episode, we begin. After each poison point, decided by the RL algorithm, we perform retraining of $\theta$ on the perturbed training set, generating $\theta^p$, and applying it to the testing set. The retraining stage is conducted by transfer learning: we take the original set of parameters and warm-start on them, and then retrain the network for a single additional epoch. 

The end-to-end system is designed to take as input a full dataset composed of graphs and have an agent learn to manipulate those graphs, using available actions in $A$, such that they maximize their episodic reward. The reward is measured as the difference between the baseline accuracy (the GNN's accuracy on the testing set before perturbation on the training set) and the perturbed accuracy (the GNN's accuracy on the testing set after the training set has been perturbed).

\section{Experiments}
\subsection{MiniGCDataset}
\begin{figure}[!ht]
\centering
\includegraphics[scale=.9]{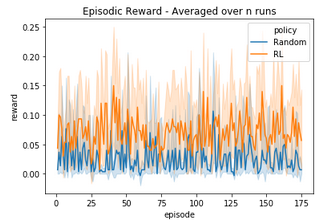}
\caption{Simulation Results on MiniGCDataset: |Train| = 180, |Test| = 30, GNN epochs = 70, RL epochs = 175. The blue line represents the accuracy degradation over a random policy, averaged out over 10 runs. The orange line is the accuracy degradation over REINFORCE, averaged out over $NRuns = 10$} 
\label{fig:minigc_test}
\end{figure}

We first create partitioned sets from the MiniGCDataset, with the size of the training set = 150, and the testing set = 30. We run the original GNN for 70 epochs to generate $\theta$ and our benchmark accuracy on the testing set. During the Poison attack, we allow the RL algorithm to perturb $p$ = 10 poison points during each episode. The reward is passed back intra-episode, between poison point selection, to reduce vanishing gradient issues and to encourage learning. This process is conducted 175 times as we fix $NEpisodes = 175$. We also fix $NRuns = 10$. The Graph Neural Network architecture is defined in Section \ref{section:GNN}. The RL algorithm procedure can be found in Algorithm \ref{algorithm: REINFORCE}, with the specific architecture released in our source code, \href{https://github.com/mb6xn/Thesis_Aug/tree/jake}{here}. Again, we use $\pi(s) = \frac{1}{|A|}$ as a baseline to measure RL's efficacy at this specific task. Figure \ref{fig:minigc_test} shows reward over the episode, averaged over $NRuns$ to reduce stochasticity. 

We see a modest improvement over random search, as well as lesser extremes in the incorrect direction, meaning we rarely see cases where our RL agent perturbs a set of graphs during an episode and increases the testing accuracy.

\begin{figure}[!ht]
\centering
\includegraphics[scale=.5]{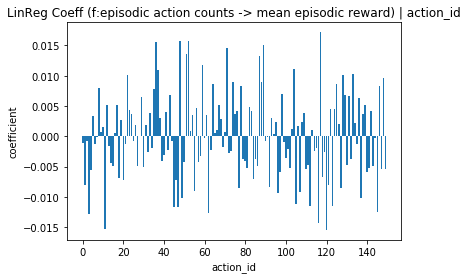}
\caption{Linear Regression: f: summed episodic actions counts -> mean episodic reward. Visualizing the coefficients to show which action selections have a higher impact on accuracy degradation over the testing set. the x-axis represents the graph id of the graph being perturbed, and the y-axis is the linear regression coefficient. *The linear regression model was fit with an $r^2$ of 0.85.}
\label{fig:minigc_linreg}
\end{figure}

We also create a way to visualize the efficacy of individual graphs perturbations in Figure \ref{fig:minigc_linreg}. There, the x-axis represents the action id, i.e., the graph id within the training set. The y-axis is the linear regression coefficient found when mapping the summed episodic actions counts to the mean episodic reward over $Nruns$. We see that star graphs appear most often amongst the top ten high valued coefficients and perceive that they are the most likely class within this dataset to have an attack result in net positive reward on the system.

\subsection{MiniGC - Larger}
\begin{figure}[!ht]
\centering
\includegraphics[scale=.75]{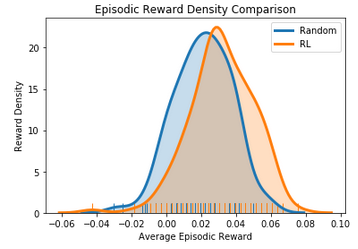}
\caption{Simulation Results on MiniGCDataset: |Train| = 180, |Test| = 30, GNN epochs = 70, RL epochs = 175. We increase the size of the networks by an order of magnitude, such that $MinNodes = 1500$ and $MaxNodes = 2000$. These are normalized density plots. The x-axis represents the average episodic reward over each policy, while the y-axis is the density of observing that statistic. Blue corresponds to a random policy, while orange is the utilized REINFORCE algorithm.} 
\label{fig:minigc_test_v2}
\end{figure}

Next, we increase the size of the graphs by an order of magnitude. All other parameters are kept the same, including the size of the subgraph inserted into the original graph. In the GNN initial training stage, we see that the size of the graphs has a profound effect on test-time accuracy, meaning higher baselines for our RL algorithms. We visualize these results in Figure \ref{fig:minigc_test_v2}, noting that the distribution plots of episodic reward comparing REINFORCE and random selection show that REINFORCE has a laterally shifted distribution and a higher peak. What we do notice is that central tendency measures are dramatically shifted downward, meaning that the increase in network sizes allowed for the GNN to draw more clear decision boundaries and consequently increase its own robustness.

\begin{figure}[!ht]
\centering
\includegraphics[scale=.75]{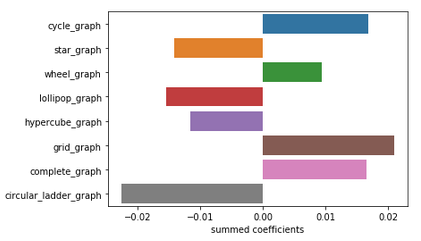}
\caption{We show the results on MiniGCDataset - Larger here, summing over the coefficients grouped by each graph class label. With the enlarged graph sizes, we see that cycle, wheel, grid and complete graphs have net positive coefficient values, which signifies relationships between the decision boundaries being draw by the GNN, as the overall structure of these four graph types makes it difficult for the model to distinguish them post-attack.}
\label{fig:minigc_linreg_2}
\end{figure}

\section{Conclusion}
In conclusion, we present a novel end-to-end poisoning attack on a GNN Classifier. The results from our analysis show that our method presents a lift over a random, brute force search of the graph space, and provides additional insight into the vulnerabilities of specific graph structures, e.g., which graph classes have a higher propensity to affect test-time performance given a perturbation. We also experiment with various ways of representing the state of our graph dynamic system. Such research could be important in practice, particularly in the healthcare industry, and has unexplored theoretical ties to the robustness and expressive power of Graph Deep Learning.

\section{Future Work}
\subsubsection{\textbf{State Representation}}
We intend to explore further methods for encoding the state of the graphical system at $t$. Here, we use a flattened representation of summary statistics over each graph in the training set, but we believe that there may be other descriptive methods from the area of network science, such as centrality measures, that may better represent a graph's characteristics in a low-dimensional space.

\subsubsection{\textbf{Action Space}}
One of the main things that we intend to explore in the future is the action space. We reduced the action space from containing granular level actions, such as node or edge edits, but believe that translating this to a Hierarchical Reinforcement Learning problem opens the door for us to operate on a larger action space without sacrificing much in the way of compute. In this way, we could subject the RL algorithm to a further customized reward function that rewards more heavily when they choose to alter the graphs in a small way, such that perturbations would be obfuscated from a human detector. Initially, our action space resembled Table \ref{tab:action}.

\begin{table}[!ht]
 \begin{tabular}{|p{1cm}||p{2cm}||p{4cm}|}
 \hline
 Action & Name & Definition  \\ [0.5ex] 
 \hline\hline
 $a_1$ & subgraph add & A random gnp graph is inserted into the existing graph structure. \\ 
 \hline
 $a_2$ & node delete & a random node with $g$ is removed \\
 \hline
 $a_3$ & node add & a node is added to $g$ \\
 \hline
 $a_4$ & edge delete & a random edge is deleted from $g$  \\
 \hline
 $a_5$ & edge add & a random edge is added to $g$\\ [1ex] 
 \hline
\end{tabular}
\caption{\label{tab:action}A hypothesized action space expansion.}
\end{table}

\subsubsection{\textbf{Baselines}}
The bulk of our work was completely experimental, and as such, there lack academically-grounded baselines as to which algorithm would perform better, or what that would mean for our results. In the future, we would like to add to our work an increasing variety of 'flat' RL methods to validate our choice, or to find a new winner. We would also like to explore the use of pretrained models, where applicable. 

\subsubsection{\textbf{Datasets}}
Most of our experimental process was centered around using the MiniGCDataset from DeepGraphLibrary. Other datasets available feature vastly different APIs, and require more work to integrate into our end to end attack. For future work, we intend to expand the testing suite to include real-world datasets e.g PROTEINS (with known consequences faced by a successful attack), and to make our overall pipeline more extensible.

\subsubsection{\textbf{Compute}}
All work noted above was completed on local machines. Some bottlenecks were seen, particularly in the GNN training/retraining stage that could benefit from parallel/distributed computing across nodes. For real-world datasets, with thousands to millions of nodes and edges, this is a worthwhile endeavor. For experimental testing on a synthetic dataset, there was not an explicit need for such support.

\subsubsection{\textbf{Hierarchical Reinforcement Learning (HRL)}}
We propose an algorithmic conversion on the RL side of our Poison attack, trading a flat method for a hierarchical one. HRL splits an  RL problem into a hierarchy of subproblems in a way that higher-level problems use lower-level problems as subroutine~\cite{Hengst2010}.  It is similar to the optimal subproblem property found in dynamic programming problems or possibly similar to divide and conquer methods; and similar to those, each of the subproblems can be reinforcement learning problems on their own. Usually, the lower-level problems only require a short sequence of actions. We can use parallel processing if the subproblems are non-overlapping, which means that HRL can reduce computational complexity, and allow for the agent to have granular control over the graph perturbation process extending beyond our work here. But parallel learning is hard since changes in a policy at one level may cause changes in higher levels. Hierarchical Actor-Critic (HAC)~\cite{levy2017learning} claims to overcome this. In HAC each level of the
hierarchy is trained independently of the lower levels by treating as if the lower level policies are already optimal. HAC has been shown to learn 3-level hierarchies in continuous state and action spaces in parallel, and is our algorithm of choice as we move toward a hierarchical approach.

{\small
\bibliographystyle{IEEEtran}
\bibliography{main}
}

\newpage
\begin{algorithm}[h!]
\caption{Poison attack($G, \pi, T, p$)}     
        $X, Y = G$   \tcp*{Extract graph label pairs}
        $f = GNN(X, Y)$   \tcp*{Train Model on X, Y training set}
        $t = 0$

    \While{$t \not= T$}   {
        //Outer loop-begin episode\\
        $t = t + 1$
        
        $poison = 0$

        $X' = X;\ \ f' = f$   \tcp*{Clone graphs and model}

        $s$ = state representations of $X$
            
        $episode\_reward =  0   $
        
        $acc\_prev$ = validation accuracy of $f$

        \While{$poison \not= p$}{

        //Inner Loop\\
                 
        $poison = poison + 1$
        
        Choose action $a$ by policy
        
        $X'  = a * X$ \tcp*{Perturb graphs with $a$}

        $s'$ = state representations of $X'$
        
        $f' = f'$ retrained one epoch on $X'$
        
        $acc\_after$ = validation accuracy of $X'$
        
        $reward$ = $acc\_prev$ - $acc\_after$
        
        $episode\_reward$ = $episode\_reward$ + $reward$
    }
}
\label{algo : 1}
\end{algorithm}

\end{document}